\documentclass{article}
\usepackage{spconf,amsmath,graphicx}
\usepackage{multirow}
\usepackage{ctable}
\usepackage{subcaption}
\usepackage{diagbox}
\usepackage{array}

\makeatletter
\newcommand{\thickhline}{%
    \noalign {\ifnum 0=`}\fi \hrule height 1pt
    \futurelet \reserved@a \@xhline
}
\newcolumntype{"}{@{\hskip\tabcolsep\vrule width 1pt\hskip\tabcolsep}}
\makeatother

\definecolor{myred}{RGB}{255, 0, 0}
\definecolor{myblue}{RGB}{70, 114, 196}
\definecolor{mygreen}{RGB}{0, 176, 80}

\title{S-DOD-CNN: Doubly Injecting Spatially-Preserved Object Information for Event Recognition}
\name{Hyungtae Lee$^{\star\dagger}$~~~~~~~~~~~~~~~~~~~~~~~~Sungmin Eum$^{\star\dagger}$~~~~~~~~~~~~~~~~~~~~~~~~Heesung Kwon$^{\star}$}
\address{$^{\star}$US Army Research Laboratory~~~~~~~~~~~~~~~~~~~~~~~~$^{\dagger}$Booz Allen Hamilton Inc.\thanks{Copyright 2020 IEEE. Published in the IEEE 2020 International Conference on Acoustics, Speech, and Signal Processing (ICASSP 2020), scheduled for 4-9 May, 2020, in Barcelona, Spain. Personal use of this material is permitted. However, permission to reprint/republish this material for advertising or promotional purposes or for creating new collective works for resale or redistribution to servers or lists, or to reuse any copyrighted component of this work in other works, must be obtained from the IEEE. Contact: Manager, Copyrights and Permissions / IEEE Service Center / 445 Hoes Lane / P.O. Box 1331 / Piscataway, NJ 08855-1331, USA. Telephone: + Intl. 908-562-3966.}
}
\begin{document}
%
\maketitle
\begin{abstract}
We present a novel event recognition approach called Spatially-preserved Doubly-injected Object Detection CNN (S-DOD-CNN), which incorporates the \emph{spatially preserved} object detection information in both a direct and an indirect way. Indirect injection is carried out by simply sharing the weights between the object detection modules and the event recognition module. Meanwhile, our novelty lies in the fact that we have preserved the spatial information for the direct injection. Once multiple regions-of-intereset (RoIs) are acquired, their feature maps are computed and then projected onto a spatially-preserving combined feature map using one of the four \emph{RoI Projection} approaches we present. In our architecture, combined feature maps are generated for object detection which are directly injected to the event recognition module. Our method provides the state-of-the-art accuracy for malicious event recognition.
\end{abstract}
\section{Introduction}
\label{sec:intro}

Object information provides crucial evidence for identifying the events shown in still images. There have been several attempts which make use of the object information in improving event recognition performance. Most methods perform event recognition with the aid of object detection results via feature-level fusion~\cite{LWangCVPRW2015,LWangICCVW2015} or score-level fusion~\cite{LLiICCV2007,TAlthoffACMMM2012,RRobinsonIROS2015,MJainCVPR2015,HLeeIROS2016,HLeeDCS2016,HLeeWACV2016,HLeeICIP2017,YCaoICIP2017,HLeeICASSP2018,HLeeTPAMI2019}.

Recently, Lee et al.~\cite{HLeeICASSP2019} introduced Doubly-injected Object Detection CNN (DOD-CNN) that incorporates the use of object detection information in a direct and an indirect way within a CNN architecture for the task of event recognition. DOD-CNN consists of three connected networks responsible for event recognition, rigid object detection, and non-rigid object detection. Three networks are co-trained while object detection information is indirectly passed onto event recognition via the shared portion of the architecture. 

DOD-CNN achieves further performance improvement by directly passing intermediate output of the rigid and non-rigid object detection onto the event recognition module. More specifically, each of the two feature maps from rigid and non-rigid object detection is generated by pooling multiple per-RoI feature maps (i.e., feature maps for each region-of-interest) via batch pooling. The two feature maps are then directly injected  into the event recognition module at the end of the last convolutional layer. Note that the batch pooling simply aggregates multiple feature maps along the batch direction without considering their spatial coordinates in the original image.

\begin{figure}[t]
\begin{minipage}[b]{1.0\linewidth}
  \centering
  \centerline{\includegraphics[width=0.85\linewidth,trim=5mm 5mm 5mm 5mm,clip]{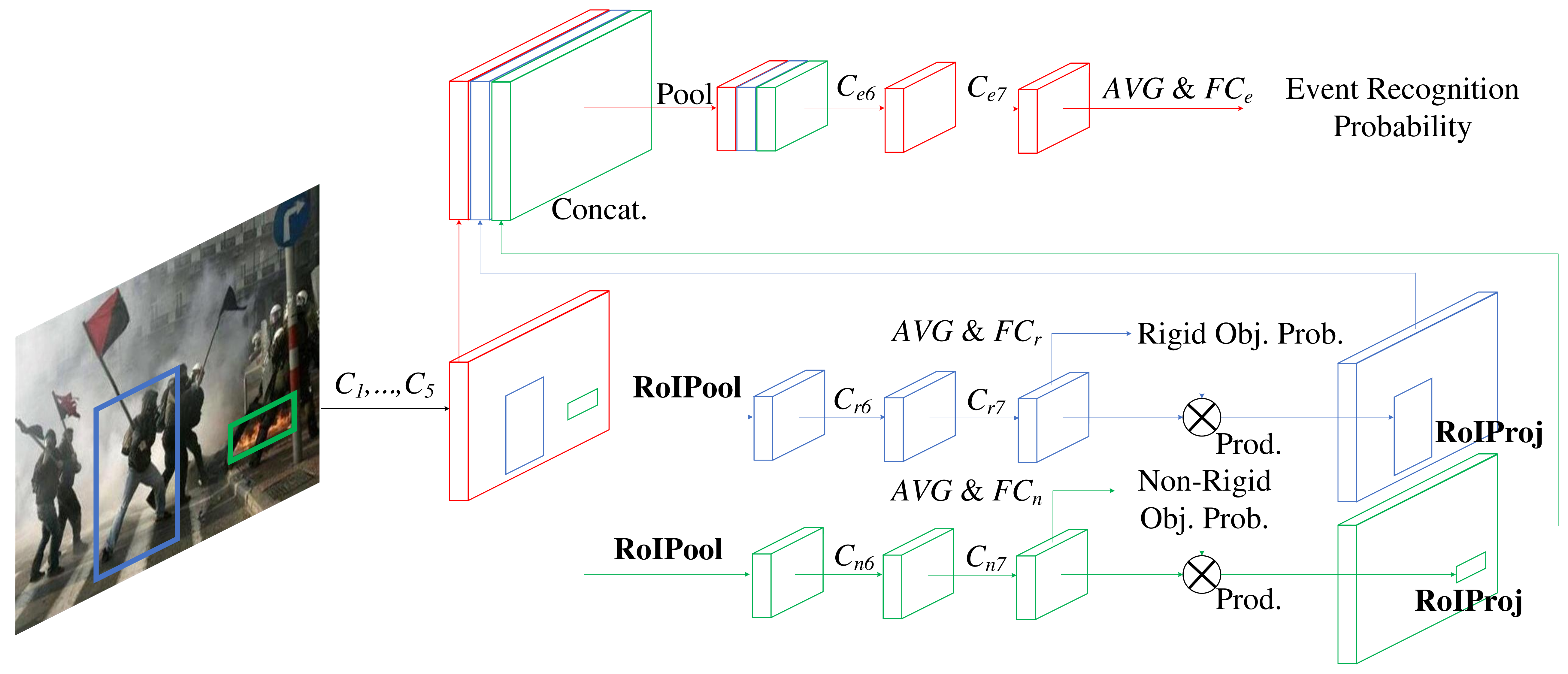}}
\end{minipage}
\vspace{-0.7cm}
\caption{{\small{\bf S-DOD-CNN Framework.} \textcolor{myred}{Red}, \textcolor{myblue}{blue}, and \textcolor{mygreen}{green} arrows indicate the computational flow responsible for event recognition ($e$), rigid object detection ($r$), and non-rigid object detection ($n$), respectively. For rigid and non-rigid object detection, a combined feature map is constructed by combining per-RoI feature maps while preserving the spatial locations of the RoIs within the original image.}}
\label{fig:architecture}
\end{figure}

In this paper, we present an approach to generate a single combined feature map which safely preserves the original spatial location of the per-RoI feature maps provided by the object detection process. Per-RoI feature maps are first projected onto separate projected feature maps using a novel \emph{RoI Projection} which are then aggregated into a single combined feature map. In the RoI projection, each per-RoI feature map is weighted by its object detection probability. Although our approach follows the spirit of DOD-CNN by incorporating the object detection information in two-ways (i.e., doubly injecting), the rigid and non-rigid object detection information is used in a different way by preserving the spatial context for each of the per-RoI feature map. Therefore, we call our new architecture \emph{Spatially-Preserved and Doubly-injected Object Detection CNN (S-DOD-CNN)} which is depicted in Figure~\ref{fig:architecture}.


When projecting the per-RoI feature maps into one single projected feature map, we adopt two interpolation methods which are \emph{MAX interpolation} and \emph{Linear interpolation}. In \emph{MAX interpolation}, a maximum value among the input points is projected into the output point. In \emph{Linear interpolation}, a linearly interpolated value of the four nearest input points is projected into the output. These interpolation methods can be applied with either \emph{class-specific} or \emph{class-agnostic} RoI selection. While class-specific selection carries out the RoI projection for each set of object class, the class-agnostic selection considers only a small subset of RoIs among all the RoIs disregarding the object classes. Therefore, the RoI projection can be conducted in four different combinations.


In order to prove the effectiveness of using a spatially-preserved object detection feature maps for event recognition, we conducted several experiments on the malicious event classification~\cite{HLeeICASSP2018}. We have validated that all four combinations of the novel RoI projection within S-DOD-CNN provide higher accuracy than all the baselines.

\section{S-DOD-CNN}
\label{sec:ourapproach}

\subsection{Architecture}
\label{ssec:architecture}

\noindent{\bf DOD-CNN.} DOD-CNN~\cite{HLeeICASSP2019} consists of five shared convolutional layers ($C_1,\cdots,C_5$), one RoI pooling layer, and three separate modules, each responsible for event recognition, rigid object detection, and non-rigid object detection, respectively. Each module consists of two convolutional layers ($C_6,~C_7$), one average pooling layer ($AVG$), and one fully connected layer ($FC$), where the output dimension of the last layer is set to match the number of events or objects.

DOD-CNN takes one image and multiple RoIs (approximatedly 2000 for rigid objects and 5 for non-rigid objects per image) as input. Selective search~\cite{JUijlingsIJCV2013} and multi-scale sliding windows~\cite{PViolaCVPR2001,NDalalCVPR2005,PFelzenszwalbTPAMI2010,HLeeACCV2012} are used to generate the RoIs for rigid and non-rigid objects, respectively. For each RoI, per-RoI feature map is computed via RoI pooling and then fed into its corresponding task-specific module.

For rigid or non-rigid object detection, the output of the last convolutional layer (denoted as \emph{per-RoI $C_7$ feature map}) is pooled into a single map along the batch direction, which is referred to as a \emph{batch pooling}. The two single feature maps are then concatenated with the output of the last convolution layer of the event recognition. The concatenated map is fed into the remaining event recognition layers which are the average pooling and fully connected layer.

Batch pooling does not preserve the spatial information of the feature maps since these maps are aligned and pooled without the consideration of their spatial coordinates in the original input image. For instance, consider selecting feature points at a same location, from two different feature maps which are aligned for batch pooling. These points do not necessarily correspond to the same location in the input image as each feature map is tied with a different RoI.
\\

\noindent{\bf S-DOD-CNN.} We introduce a novel method that aggregates multiple feature maps which come from different regions in the input image while preserving the spatial information. The spatial information for each per-RoI $C_7$ feature map is preserved by projecting each feature map onto a location on a \emph{projected feature map} which corresponds to its original spatial location within the input image. 

\begin{figure}[t]
\begin{minipage}[b]{1.0\linewidth}
  \centering
  \centerline{\includegraphics[width=0.8\linewidth,trim=5mm 5mm 5mm 5mm,clip]{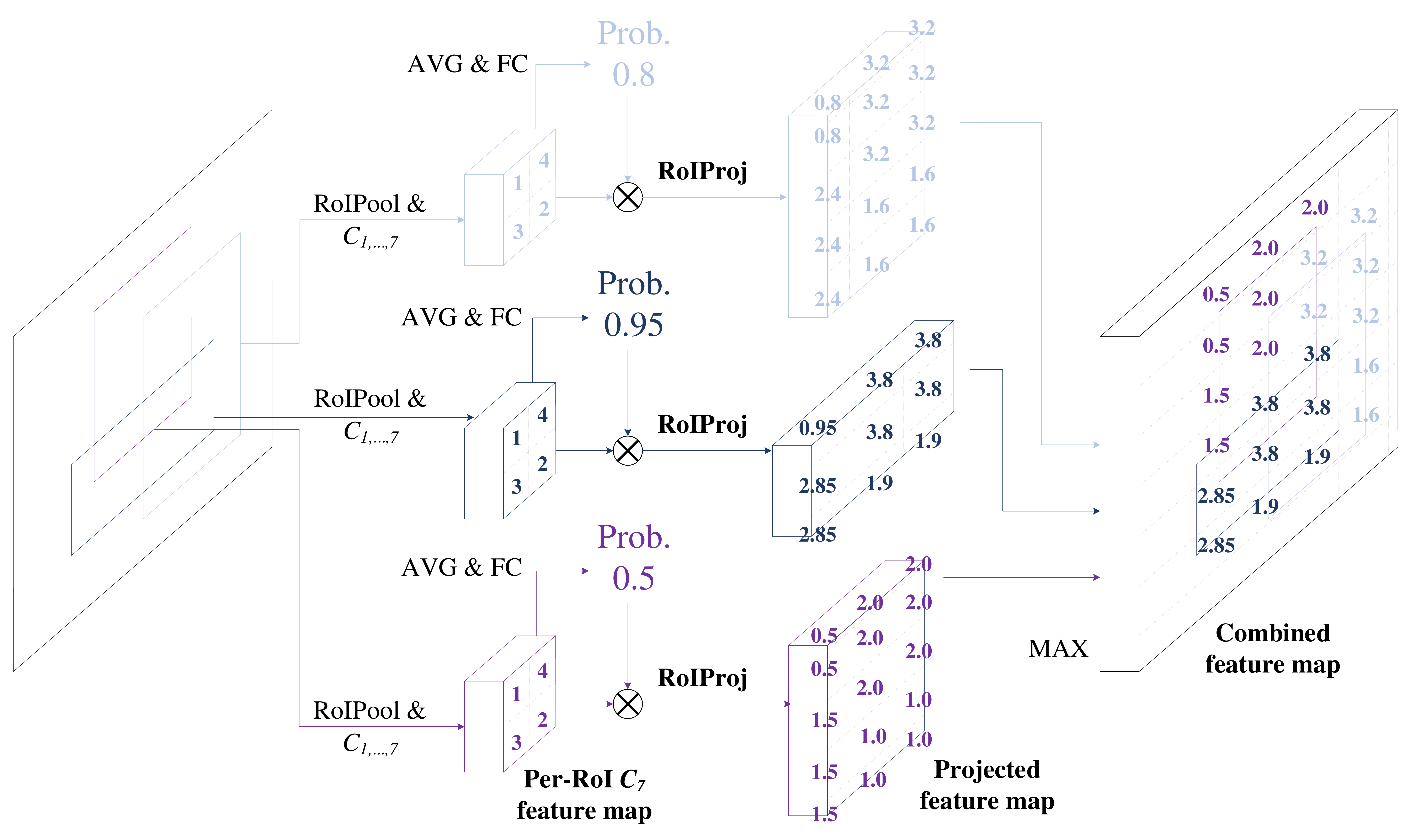}}
\end{minipage}
\vspace{-0.7cm}
\caption{{\small{\bf Overall Process of Building a Combined Feature Map.} The combined feature map is max-pooled with multiple projected feature maps that are projected from original feature maps (2$\times$2 bins in this example) w.r.t. their original spatial coordinates in the image.}}
\label{fig:map_building}
\end{figure}

Figure~\ref{fig:map_building} illustrates how per-RoI $C_7$ feature maps are processed through RoI Projection (\emph{RoIProj}) to generate corresponding projected feature maps. Note that before the per-RoI $C_7$ feature maps are fed into RoIProj, they are multiplied by its detection probability to incorporate the reliability for each detection result. The projected feature maps are then max-pooled to build a \emph{combined feature map}. In our experiment, five per-RoI $C_7$ feature maps with the highest probability values are chosen to build the combined feature map.

\begin{figure}[t]
\begin{minipage}[b]{0.475\linewidth}
  \centering
  \centerline{\includegraphics[width=0.8\linewidth,trim=5mm 5mm 5mm 5mm,clip]{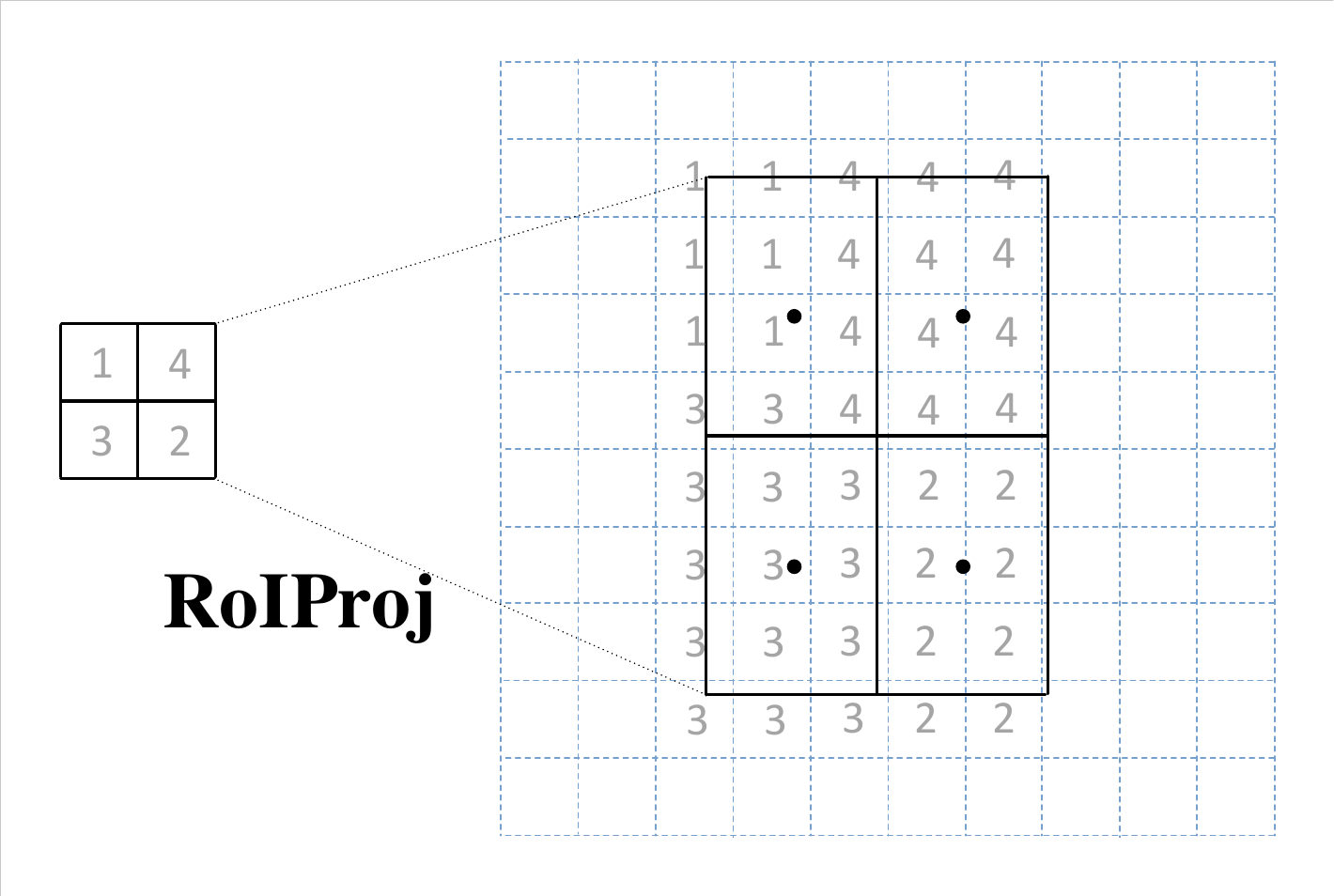}}
  \subcaption{{\small MAX Projection}}
\end{minipage}~~~~~~
\begin{minipage}[b]{0.475\linewidth}
  \centering
  \centerline{\includegraphics[width=0.8\linewidth,trim=5mm 5mm 5mm 5mm,clip]{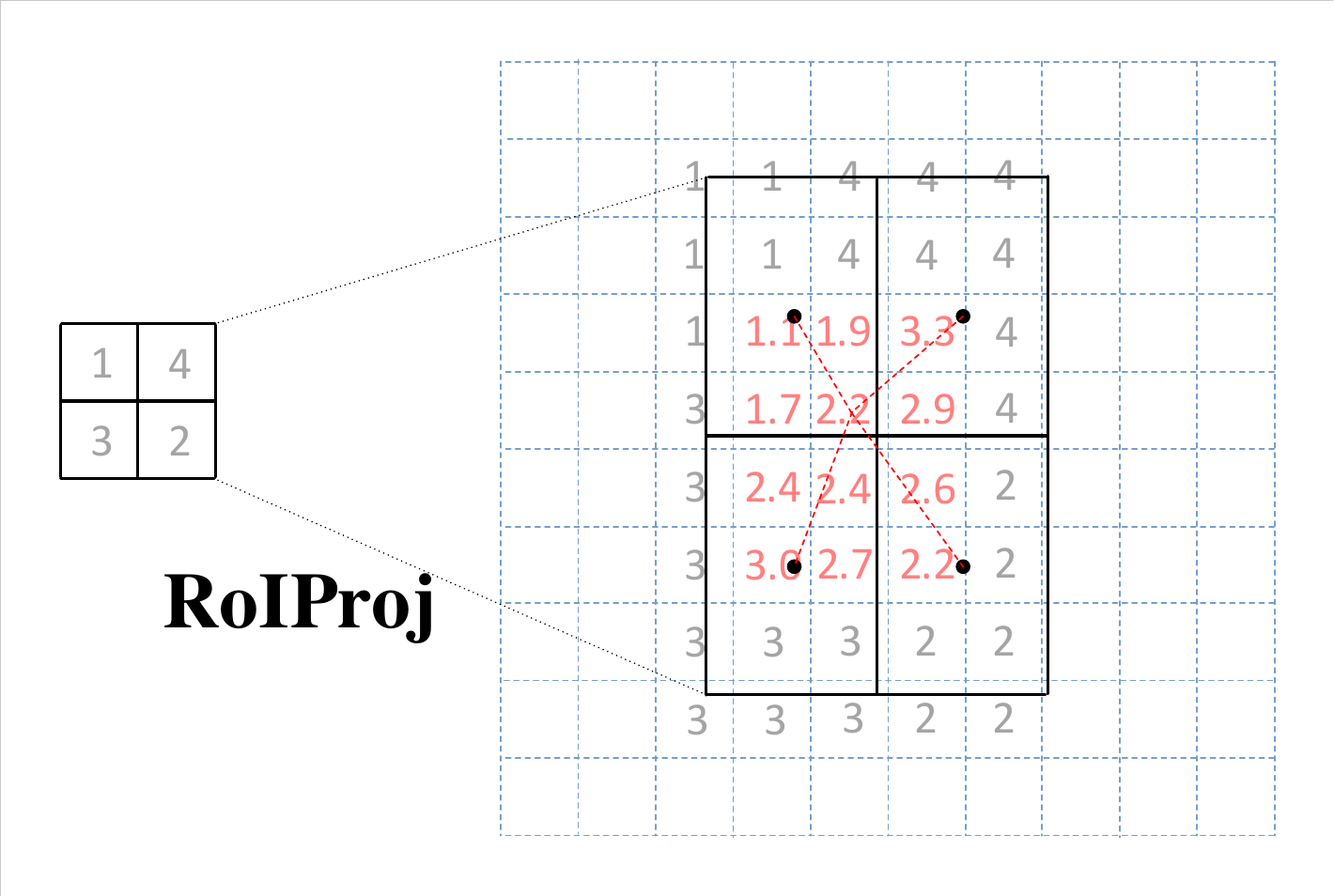}}
  \subcaption{{\small Linear Projection}}
\end{minipage}
\vspace{-0.7cm}
\caption{{\small{\bf RoI Projection.}}}
\label{fig:RoIProj}
\end{figure}

\begin{figure*}[t]
\begin{minipage}[b]{1.0\linewidth}
  \centering
  \centerline{\includegraphics[width=0.7\linewidth,trim=5mm 5mm 5mm 5mm,clip]{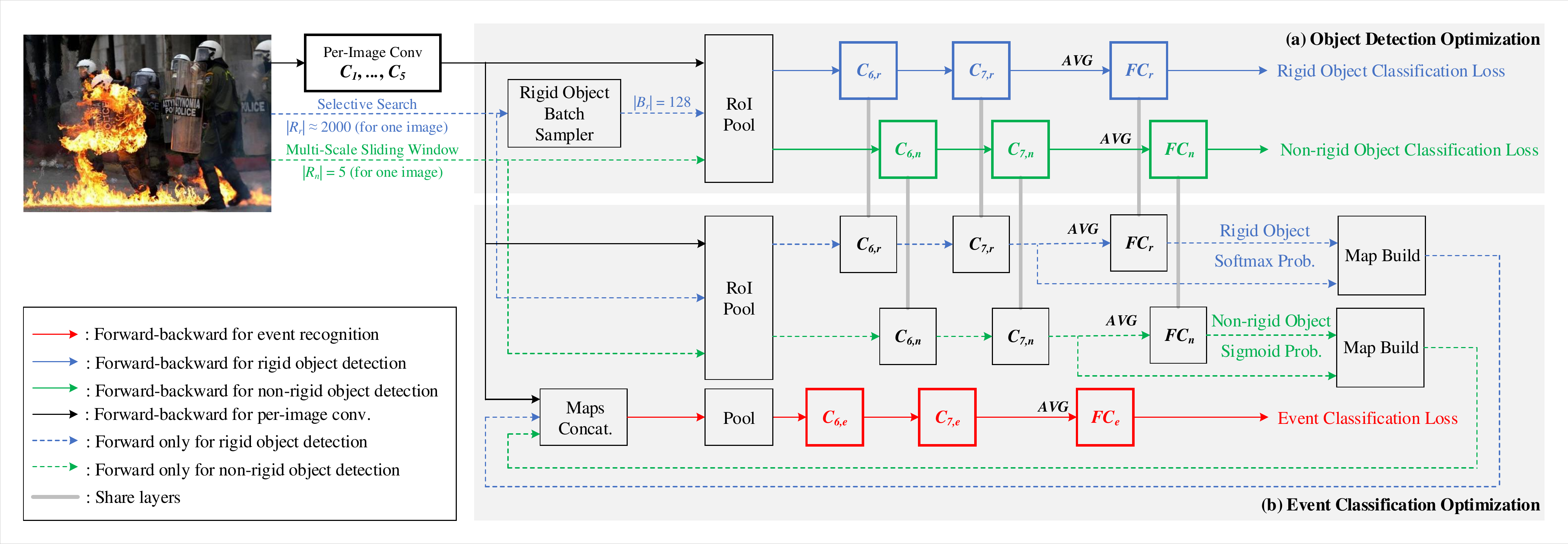}}
\end{minipage}
\vspace{-0.7cm}
\caption{{\small{\bf The Second Stage of Training Process.}}}
\label{fig:training}
\end{figure*}

Our network generates two separate combined feature maps, one for rigid and another for non-rigid object detection. These two combined feature maps are concatenated with the event recognition feature map as in Figure~\ref{fig:architecture}. The two combined feature maps share the same-sized and aligned receptive field with the event recognition feature map, and thus they are `spatially-preserved'. The event recognition feature map is the output of $C_5$ layer right before RoI pooling. The event recognition module intakes this concatenated map to compute the event recognition probability. As we are constructing our network based on the DOD-CNN, but with `spatially-preserved' object detection information for event recognition, we call it Spatially-preserved DOD-CNN (S-DOD-CNN).\\

\noindent{\bf RoI Projection.} When projecting the per-RoI $C_7$ feature maps into one projected feature map (denoted as RoIProj in Figure~\ref{fig:map_building}), we adopt one of the two interpolation methods: \emph{MAX interpolation} or \emph{Linear interpolation}. Examples of the two interpolations are shown in Figure~\ref{fig:RoIProj}. When multiple points on an input map is being projected onto a single point on an output map, the point is filled with a maximum (\emph{MAX}) or a linearly interpolated value of four nearest input points (\emph{Linear}).

The RoI projection can be performed in two different ways. These two methods differ based on how a subset of RoIs (the RoIs that are actually used for projection) is selected from the overall set of RoIs. Both of the selection methods utilize $N$ probability scores which are generated for each RoI after AVG \& FC (see Figure~\ref{fig:map_building}), where $N$ is the number of classes. For \emph{class-specific} selection, 5 RoIs with highest probability scores are chosen for each class. For \emph{class-agnostic} selection, 5 RoIs with highest probability scores are chosen from all the RoIs without regard to which classes they come from. Therefore, the number of executions for RoI projection is either $N$ times or just once, based on which selection method is chosen. In addition, if per-RoI $C_7$ feature map has $k$ channels, the dimension of the channel for the combined map under the class-specific selection becomes $k\times N$, while it remains as $k$ under the class-agnostic selection. Overall, the RoI projection can be performed as one of the four combinations as there are two different interpolations methods (MAX/Linear) and two different RoI selection methods (class-specific/agnostic).

\subsection{Training}
\label{ssec:training}

S-DOD-CNN is trained by using a mini-batch stochastic gradient descent (SGD) optimization approach. Event recognition and rigid object detection modules are optimized by minimizing their softmax loss while cross entropy loss is used for non-rigid object detection optimization. Each batch contains two images consisting of one malicious image and one benign image. For event recognition and non-rigid object detection, 1 and 5 RoIs are generated for each image, respectively. For rigid object detection, a batch takes 64 RoIs randomly selected from approximately 2000 RoIs generated by selective search. Accordingly, we need to prepare a large number of batches to cover the entire RoI set for training rigid object detection. A batch (which contains 2 images) consists of 2, 128, and 10 RoIs for event recognition, rigid object detection, and non-rigid object detection, respectively.

In preparing the positive and negative samples for training, we have used 0.5 and 0.1 as the rigid and non-rigid object detection thresholds for the intersection-over-union (IOU) metric, respectively. Any RoI whose IOU with respect to the ground truth bounding box is larger than the threshold is treated as a positive example. RoIs whose IOU is lower than 0.1 are treated as negative examples.

The weights in $C_1,\cdots,C_5$ are initially inherited from the pre-trained AlexNet~\cite{AKrizhevskyNIPS2012} trained on a large-scale Places dataset~\cite{BZhouNIPS2014} and the remaining layers ($C_6$, $C_7$ and $FC$ layers for all three modules) are initialized according to Gaussian distribution with 0 mean and 0.01 standard deviation.\\

\noindent{\bf Two-stage Cascaded Optimization.} To allow more batches for training rigid object detection, we use a two-stage cascaded optimization strategy. In the first stage, only the layers used to perform rigid object detection are trained. Then, in the second stage, all three tasks are jointly optimized in an end-to-end fashion. Figure~\ref{fig:training} shows the second stage of the training process. For each training iteration in the second stage, two processes ((a) and (b) in Figure~\ref{fig:training}) are executed in order. In process (a), all the layers of the two object detection modules are optimized with a batch containing 128 RoIs of rigid object and 10 RoIs of non-rigid object. After the process (a) is done, full set of RoIs (i.e. approximately 4000 RoIs for rigid object, 10 RoIs for non-rigid object) is fed into the object detection modules. The resulting combined feature maps are injected into the event recognition module for optimization. We set the learning rate of 0.001, 50k iterations, and the step size of 30k for the first stage and the learning rate of 0.0001, 20k iterations, and the step size of 12k for the second stage.

\section{Experiments}
\label{sec:exp}

\subsection{Dataset}
\label{ssec:dataset}

Malicious Crowd Dataset~\cite{HLeeICASSP2018,SEumDCS2018} is selected as it provides the appropriate components to evaluate the effects of using object information for event recognition. It contains 1133 images and is equally divided into {\it malicious} classes and {\it benign} classes. Half of the dataset is used for training and the rest is used for testing. In addition to the label of the event class, bounding box annotations of three rigid objects ({\it police}, {\it helmet}, {\it car}) and two non-rigid objects ({\it fire}, {\it smoke}) are provided. \cite{HLeeICASSP2018} provides details on how these objects are selected.

\subsection{Performance Evaluation}
\label{ssec:performeval}

To demonstrate the effectiveness of our approach, we compared the event recognition accuracy of S-DOD-CNN with two baselines: DOD-CNN which does not include direct injection and DOD-CNN which incorporates both direct and indirect injection. The accuracy is measured with average precision (AP) as shown in Table~\ref{tab:performance}. S-DOD-CNN, which adopts one of the four RoI projections, provides at least 1.1\% higher accuracy than both of the baselines. This verifies the effectiveness of using object detection information spatially preserved via RoI projection. RoI projection using linear interpolation and class-specific RoI selection shows the highest accuracy among all the methods but the differences are marginal.

\begin{table}[t]
\setlength{\tabcolsep}{5.0pt}
\renewcommand{\arraystretch}{0.9}
\begin{center}
{\small
\begin{tabular}{c|cc|c}
\specialrule{.15em}{.05em}{.05em}
\multirow{2}{*}{Method} & \multicolumn{2}{c|}{RoI Projection} & \multirow{2}{*}{AP (\%)} \\
 & MAX/Linear & RoI Selection & \\
\specialrule{.15em}{.05em}{.05em}
No Direct Inject.~\cite{HLeeICASSP2019} & $\cdot$ & $\cdot$ & 90.7 \\
DOD-CNN~\cite{HLeeICASSP2019} & $\cdot$ & $\cdot$ & 94.6 \\\hline
\multirow{4}{*}{S-DOD-CNN} & MAX & Class-agnostic & 95.7 \\
 & MAX & Class-specific & 95.8 \\
 & Linear & Class-agnostic & 95.8 \\
 & Linear & Class-specific & {\bf 95.9} \\
\specialrule{.15em}{.05em}{.05em}
\end{tabular}
}
\end{center}
\vspace{-0.6cm}
\caption{{\small {\bf Event recognition average precision (AP).} All networks use the same backbone consisting of five per-image convolution layers and three sets of two convolutional layers and one fully connected layer corresponding to three tasks.}}
\label{tab:performance}
\end{table}

In Table~\ref{tab:sigle_vs_multi}, we also analyzed how the each task performs when they are individually optimized (Single-task) or co-optimized. For No Direct Injection and DOD-CNN cases, non-rigid object detection performs better when optimized simultaneously with other tasks. However, in S-DOD-CNN, the performance of the two sub-tasks (rigid and non-rigid object detection) was degraded. This indicates that the two tasks are sacrificed to solely improve event recognition performance.

\begin{table}[t]
\setlength{\tabcolsep}{2.6pt}
\renewcommand{\arraystretch}{0.9}
\begin{center}
{\small
\begin{tabular}{c|c|c|c|c}
\specialrule{.15em}{.05em}{.05em}
Task & Single-task &  No Direct Injection & DOD-CNN & S-DOD-CNN \\
\specialrule{.15em}{.05em}{.05em}
E &  89.9 & 90.7 & 94.6 & 95.8 \\
R & 8.1 & 7.8 & 7.8 & 7.7 \\
N & 30.4 & 37.2 & 37.2 & 22.5 \\
\specialrule{.15em}{.05em}{.05em}
\end{tabular}
}
\end{center}
\vspace{-0.6cm}
\caption{{\small{\bf Single task versus multitask performance.} Task: {\bf E}: Event Recognition, {\bf R}: Rigid Object Detection, {\bf N}: Non-rigid Object Detection. For S-DOD-CNN, RoI Projection with linear interpolation and class-specific selection was chosen.}}
\label{tab:sigle_vs_multi}
\end{table}

\subsection{Ablation Study: Location of Building and Injecting Combined Feature Map}
\label{ssec:ablation}

Applying a convolutional layer after the concatenation may not be effective if the combined feature maps (coming from object detection) are not aligned properly with the event recognition feature map. One advantage achieved by constructing combined feature maps using our approach is that the map can be injected at any position in event recognition. Table~\ref{tab:pool_concat_position} shows the performance that varies according to the location of building and injection of the combined feature map. DOD-CNN, which loses RoI's spatial information during building a feature map, shows performance degradation when the injection location is placed before any convolutional layer (i.e., $C_6$ in Table~\ref{tab:pool_concat_position}). In contrast, S-DOD-CNN does not lose any performance regardless of the injection position.

The performance of S-DOD-CNN depends greatly on the building location of the combined feature map. The best accuracy is achieved when it is constructed after $C_7$. Letting the input image go through more number of convolutional layers before building the combined feature maps may have provided a richer representation.


\begin{table}[t]
\setlength{\tabcolsep}{9.5pt}
\renewcommand{\arraystretch}{0.9}
\begin{center}
{\small
\begin{tabular}{c|c"ccc}
\specialrule{.15em}{.05em}{.05em}
Method & \diagbox[width=7em]{Build}{Inject} & $C_5$ & $C_6$ & $C_7$ \\\specialrule{.15em}{.05em}{.05em}
DOD-CNN~\cite{HLeeICASSP2019} & $C_7$ & $\cdot$ & 91.4 & 94.6 \\\hline
\multirow{3}{*}{S-DOD-CNN} & RoIPool & 90.5 & 90.6 & 90.5 \\
& $C_6$ & 94.8 & 94.8 & 94.7 \\
& $C_7$ & {\bf 95.8} & 95.7 & 95.5 \\
\specialrule{.15em}{.05em}{.05em}
\end{tabular}
}
\end{center}
\vspace{-0.6cm}
\caption{{\small{\bf Performance comparison w.r.t. location of building and injection of combined feature maps.} RoI projection with linear interpolation and class-specific selection used for S-DOD-CNN.}}
\label{tab:pool_concat_position}
\end{table}

\section{Conclusion}
\label{sec:concl}
We have devised an event recognition approach referred to as S-DOD-CNN where the object detection is exploited while preserving the spatial information. Multiple per-RoI feature maps within an object detection module are projected onto a combined feature map using one of the newly presented RoI Projections preserving the spatial location of each RoI with respect to the original image. These maps are then injected to the event recognition module. Our approach provides the state-of-the-art accuracy for malicious event recognition.

\bibliographystyle{IEEEbib}
\bibliography{refs}

\end{document}